\documentclass[conference]{IEEEtran}
\usepackage{cite}

\ifCLASSINFOpdf
\usepackage[pdftex]{graphicx}
\else
\fi
\usepackage{amsmath}
\usepackage{amsfonts} 
\usepackage{amsthm}
\usepackage{amssymb}
\usepackage[boxruled]{algorithm2e}
\usepackage{url}

\usepackage{bbm}

\usepackage{kbordermatrix}

\newcommand{\pfun}{\mathop{\hbox{$\to$\kern-7pt\raise.9pt\hbox{\scalebox{1}[.55]{$|$}}\kern4pt} }}

\usepackage{array}

\begin{document}

\title{GraphChallenge.org \\ Sparse Deep Neural Network Performance}

\author{\IEEEauthorblockN{Jeremy Kepner$^{1,2,3}$, Simon Alford$^2$, Vijay Gadepally$^{1,2}$, \\ Michael Jones$^1$, Lauren Milechin$^4$, Albert Reuther$^1$, Ryan Robinett$^3$, Sid Samsi$^1$
\\
\IEEEauthorblockA{$^1$MIT Lincoln Laboratory Supercomputing Center, $^2$MIT Computer Science \& AI Laboratory, \\\ $^3$MIT Mathematics Department, $^4$MIT Dept. of Earth, Atmospheric, \& Planetary Sciences
}}}
\maketitle

\begin{abstract}
The MIT/IEEE/Amazon GraphChallenge.org encourages community approaches to developing new solutions for analyzing graphs and sparse data.  Sparse AI analytics present unique scalability difficulties.  The  Sparse Deep Neural Network (DNN) Challenge draws upon prior challenges from machine learning, high performance computing, and visual analytics to create a challenge that is reflective of emerging sparse AI systems.  The sparse DNN challenge is based on a mathematically well-defined DNN inference  computation and can be implemented in any programming environment.  In 2019 several sparse DNN challenge submissions were received from a wide range of authors and organizations.  This paper presents a performance analysis of the best performers of these submissions.   These submissions show that their state-of-the-art sparse DNN execution time, $T_{\rm DNN}$, is a strong function of the number of DNN operations performed, $N_{\rm op}$.  The sparse DNN challenge provides a clear picture of current sparse DNN systems and underscores the need for new innovations to achieve high performance on very large sparse DNNs.
\end{abstract}

%
\IEEEpeerreviewmaketitle

\section{Introduction}
\let\thefootnote\relax\footnotetext{This material is based upon work supported by the Assistant Secretary of Defense for Research and Engineering under Air Force Contract No. FA8702-15-D-0001 and National Science Foundation CCF-1533644. Any opinions, findings, conclusions or recommendations expressed in this material are those of the author(s) and do not necessarily reflect the views of the Assistant Secretary of Defense for Research and Engineering or the National Science Foundation.}

MIT/IEEE/Amazon GraphChallenge.org encourages community approaches to developing new solutions for analyzing graphs and sparse data.  GraphChallenge.org provides a well-defined community venue for stimulating research and highlighting innovations in graph and sparse data analysis software, hardware, algorithms, and systems. The target audiences for these challenges are individuals or teams that seek to highlight their contributions to graph and sparse data analysis software, hardware, algorithms, and/or systems.

As research in artificial neural networks progresses, the sizes of state-of-the-art deep neural network (DNN) architectures put increasing strain on the hardware needed to implement them \cite{7298594, kepner_exact}. In the interest of reduced storage and runtime costs, much research over the past decade has focused on the sparsification of artificial neural networks \cite{lecun1990optimal,hassibi1993second,srivastava2014dropout,iandola2016squeezenet,DBLP:journals/corr/SrinivasB15,DBLP:journals/corr/HanMD15,7298681,KepnerGilbert2011,kepner2017enabling,kumar2018ibm,kepner2018mathematics}. In the listed resources alone, the methodology of sparsification includes Hessian-based pruning \cite{lecun1990optimal,hassibi1993second}, Hebbian pruning  \cite{srivastava2014dropout}, matrix decomposition \cite{7298681}, and graph techniques \cite{kumar2018ibm,KepnerGilbert2011,kepner2017enabling,kepner2018mathematics}.
The sparse DNN challenge seeks to highlight innovations that are applicable to emerging sparse AI and machine learning \cite{SparseDNNchallenge}. 

Challenges such as YOHO~\cite{yoho}, MNIST~\cite{mnist}, HPC Challenge~\cite{hpcc}, ImageNet~\cite{imagenet} and VAST~\cite{vast1,vast2} have played important roles in driving progress in fields as diverse as machine learning, high performance computing and visual analytics. YOHO is the Linguistic Data Consortium database for voice verification systems and has been a critical enabler of speech research. The MNIST database of handwritten letters has been a bedrock of the computer vision research community for two decades. HPC Challenge has been used by the supercomputing community to benchmark and acceptance test the largest systems in the world as well as stimulate research on the new parallel programing environments. ImageNet populated an image dataset according to the WordNet hierarchy consisting of over 100,000 meaningful concepts (called synonym sets or synsets)~\cite{imagenet} with an average of 1000 images per synset and has become a critical enabler of vision research. The VAST Challenge is an annual visual analytics challenge that has been held every year since 2006; each year, VAST offers a new topic and submissions are processed like conference papers.  The sparse DNN  challenge seeks to draw on the best of these challenges, but particularly the VAST Challenge in order to highlight innovations across the algorithms, software, hardware, and systems spectrum.

The focus on graph analytics allows the sparse DNN challenge to also
draw upon significant work from the graph benchmarking community. Scale is an important driver of the Graph Challenge and graphs with billions to trillions of edges are of keen interest.  The Graph Challenge is designed to work on arbitrary graphs drawn from both real-world data sets and simulated data sets.
Examples of real-world data sets include the Stanford Large Network Dataset
Collection (see http://snap.stanford.edu/data), the AWS Public Data Sets (see
aws.amazon.com/public-data-sets), and the Yahoo! Webscope Datasets (see
webscope.sandbox.yahoo.com).  These real-world data sets cover a wide range of
applications and data sizes.  While real-world data sets have many contextual
benefits, synthetic data sets allow the largest possible graphs to be readily
generated. Examples of synthetic data sets include Graph500, Block Two-level
Erdos-Renyi graph model (BTER) \cite{seshadhri2012community}, Kronecker Graphs
\cite{KepnerGilbert2011,sanders2018triangle,kepner2019radixnet}, and  Perfect Power Law graphs
\cite{kepner2012perfect,gadepally2015using,kepner2018powerlaw}. The focus of the Graph Challenge is on graph analytics.  While parsing and formatting
complex graph data are necessary in any graph analysis system, these data sets are
made available to the community in a variety of pre-parsed formats to minimize the
amount of parsing and formatting  required by Graph Challenge participants.  The public data
are available in a variety of formats, such as linked list, tab separated, and
labeled/unlabeled.

The Graph Challenge consists of a pre-challenge and three full challenges
\begin{itemize}
\item Pre-challenge: PageRank pipeline \cite{dreher2016pagerank}
\item Static graph challenge: subgraph isomorphism \cite{samsi2017static}
\item Streaming graph challenge: stochastic block partition \cite{ed}
\item Sparse DNN challenge \cite{SparseDNNchallenge}
\end{itemize}
The static graph challenge is further broken down into triangle counting and k-truss.  The sparse DNN challenge is the focus of this paper. The organization of this paper is as follow.  First, a recap of the sparse DNN challenge is provided.  Next, an overview is presented of the 2019 submissions.  The core of the paper is the section on the analysis of the  submissions that performed sparse DNN challenge.  Based on this analysis, these results are synthesized to provide a picture of the current state-of-the-art.

\section{Deep Neural Networks}

Machine learning has been the foundation of artificial intelligence since its inception
\cite{ware1955introduction,clark1955generalization,selfridge1955pattern,dinneen1955programming,newell1955chess,mccarthy2006proposal,minsky1960learning,minsky1961steps}. Standard machine learning applications include speech recognition \cite{selfridge1955pattern}, computer vision \cite{dinneen1955programming}, and even board games \cite{newell1955chess,samuel1959some}.

\begin{figure}[htb]
  	\centering
    	\includegraphics[width=\columnwidth]{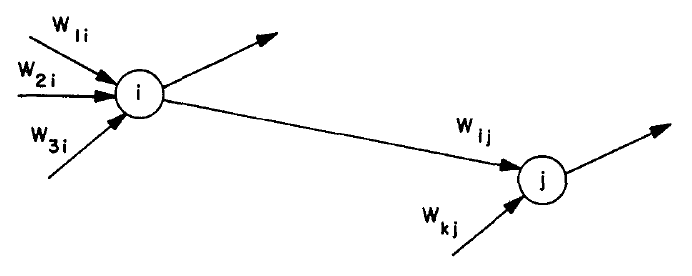}
      	\caption{Typical network elements $i$ and $j$ showing connection weights $w$ (reproduced from  \cite{clark1955generalization})}
      	\label{fig:clark1955fig1}
\end{figure}

\begin{figure}[htb]
  	\centering
    	\includegraphics[width=\columnwidth]{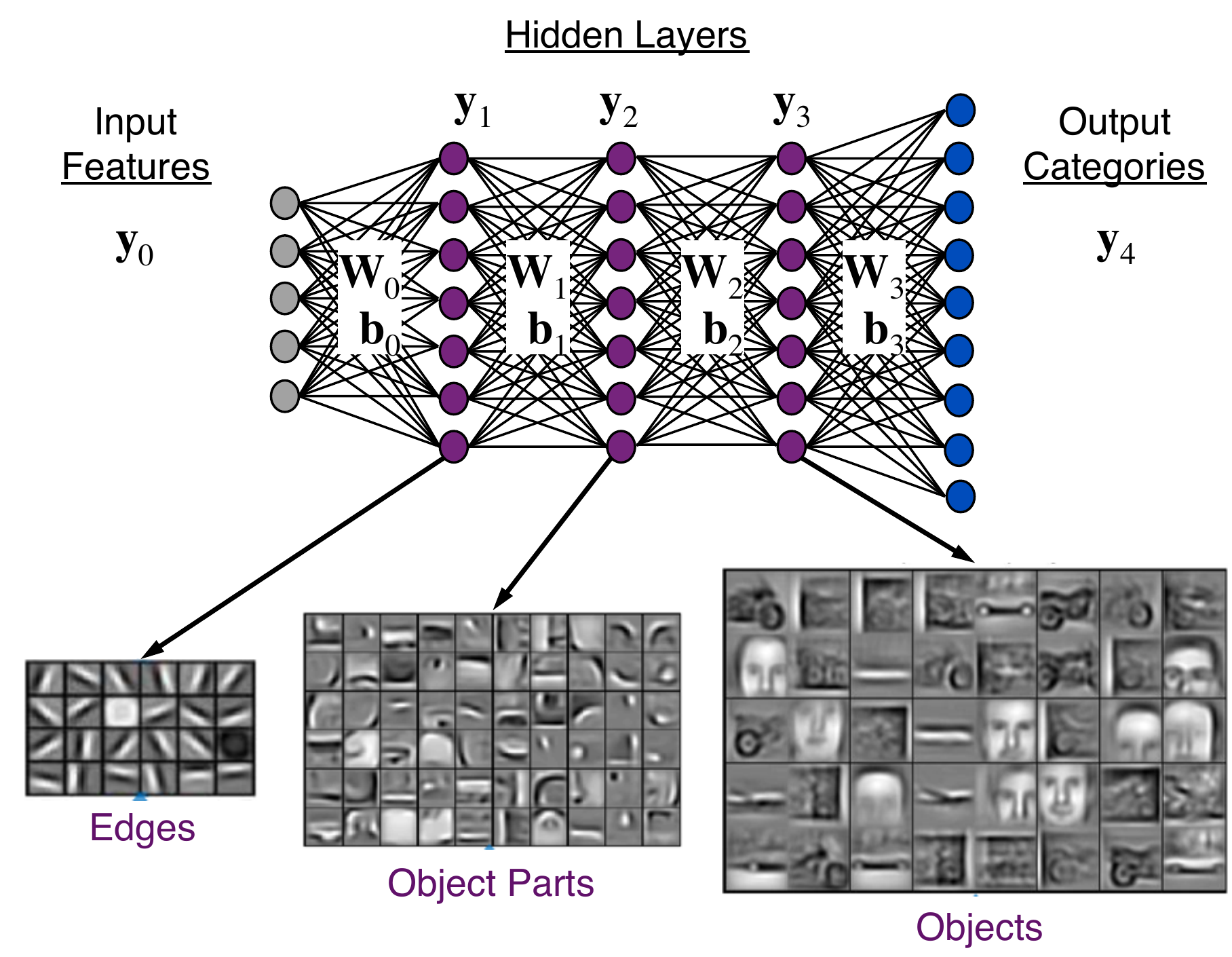}
	\caption{Four layer ($L=4$) deep neural network architecture
	for categorizing images.  The input
	features  ${\bf y}_0$ of an image are passed through a series
	of network layers ${\bf W}_{\ell=0,1,2,3}$, with bias terms
	${\bf b}_{\ell=0,1,2,3}$, that produce scores for categories
	${\bf y}_{L=4}$.  (Figure adapted from \cite{lee2009convolutional})}
      	\label{fig:DNNarchitecture}
\end{figure}

Drawing inspiration from biological neurons to implement machine learning was the topic of the first paper presented at the first machine learning conference in 1955 \cite{ware1955introduction,clark1955generalization} (see Figure~\ref{fig:clark1955fig1}). It was recognized very early on in the field that direct computational training of neural networks was computationally unfeasible with the computers that were available at that time \cite{minsky1960learning}.  The many-fold improvement in neural network computation and theory has made it possible to create neural networks capable of better-than-human performance in a variety of domains \cite{lippmann1987introduction,reynolds2000speaker,krizhevsky2012imagenet,lecun2015deep}. The production of validated data sets \cite{campbell1995testing,lecun1998mnist,deng2009imagenet} and the power of graphic processing units (GPUs) \cite{campbell2002deep,mcgraw2007benchmarking,kerr2008gpu,epstein2012making}
have allowed the effective training of deep neural networks (DNNs) with 100,000s of input features, $N$, and 100s of layers, $L$, that are capable of choosing from among 100,000s categories, $M$ (see Figure~\ref{fig:DNNarchitecture}).

The impressive performance of large DNNs provides motivation to explore even larger networks.  However, increasing $N$, $L$, and $M$ each by a factor 10 results in a 1000-fold increase in the memory required for a DNN.  Because of these memory constraints, trade-offs are currently being made in terms of precision and accuracy to save storage and computation \cite{liu2015sparse,lavin2016fast,jouppi2017datacenter,kepner2017enabling}. Thus, there is significant interest in exploring the effectiveness of sparse DNN representations where many of the weight values are zero.  As a comparison, the human brain has approximately 86 billion neurons and 150 trillion synapses~\cite{CNE:CNE21974}.  Its graph representation would have approximately 2,000 edges per node, or a density of $2 \times 10^3 / 86 \times 10^9 = 0.000002\%$.

If a large fraction of the DNN weights can be set to zero, storage and computation costs can be reduced proportionately \cite{iandola2016squeezenet,shi2017speeding}.  The interest in sparse DNNs is not limited to their computational advantages. There has also been extensive theoretical work exploring the potential neuromorphic and algorithmic benefits of sparsity \cite{lee2008sparse,boureau2008sparse,glorot2011deep,DBLP:journals/corr/HanMD15,yu2012exploiting}.

  The primary mathematical operation performed by a DNN network is the inference, or forward propagation, step.  Inference is executed repeatedly during training to determine both the weight matrix ${\bf W}_\ell$ and the bias vectors ${\bf b}_\ell$ of the DNN.  The inference computation shown in Figure~\ref{fig:DNNarchitecture} is given by
$$
  {\bf y}_{\ell + 1} = h({\bf y}_\ell {\bf W}_\ell  + {\bf b}_\ell)
$$
where $h()$ is a nonlinear function applied to each element of the vector.  The sparse DNN challenge uses the standard graph community convention whereby ${\bf W}(i,j) \neq 0$  implies a connection between neuron $i$ and neuron $j$.  In this convention ${\bf y}_\ell$ are row vectors and left matrix multiply  is used to progress through the network.  Standard AI definitions can be used  by transposing all matrices and multiplying on the  right.
A commonly used function is the rectified linear unit (ReLU) given by
$$
   h({\bf y}) = \max({\bf y},0)
$$
which sets values less that 0 to 0 and leaves other values unchanged.  For the Sparse DNN challenge, $h()$ also has an upper limit set to 32. When training a DNN, or performing inference on many different inputs, it is usually necessary to compute multiple ${\bf y}_\ell$ vectors at once in a batch that can be denoted as the matrix ${\bf Y}_\ell$.  In matrix form, the inference step becomes
$$
  {\bf Y}_{\ell + 1} = h({\bf Y}_\ell {\bf W}_\ell  + {\bf B}_\ell)
$$
where ${\bf B}_\ell$ is a replication of ${\bf b}_\ell$ along columns given by
$$
  \mathbf{B}_\ell = \mathbf{b}_\ell |\mathbf{Y}_\ell \mathbf{1}|_0
$$
and $\mathbf{1}$ is a column array of 1's, and $| ~ |_0$ is the zero norm.

\section{Neural Network Data}

Scale is an important driver of the Graph Challenge and graphs with billions to trillions of edges are of keen interest. Real sparse neural networks of this size are difficult to obtain from real data.  Until such data is available, a reasonable first step is to simulate data with the desired network properties with an emphasis on the difficult part of the problem, in this case: large sparse DNNs. The RadiX-Net synthetic sparse DNN generator is used \cite{robinett2019radix-net} to efficiently generate a wide range of pre-determined DNNs all with 32 connections per neuron.  RadiX-Net produces DNNs with a number of desirable properties, such as equal number of paths between all inputs, outputs, and intermediate layers.  The RadiX-Net DNN generation algorithm uses mixed radices to generate DNNs of specified connectedness which are then expanded via Kronecker products into larger DNNs. The number of connections (see Table~\ref{tab:RadiX-Net-DNNs}) in the resulting large sparse DNNs are computed using the formula  
$$
   N_{\rm c} = 32 \times L \times N
$$

\begin{table}
\centering
\begin{tabular}{ccccc}
\hline
                & \textbf{Neurons} & \textbf{Neurons} & \textbf{Neurons} & \textbf{Neurons}\\
\textbf{Layers} & 1024             & 4096             & 16384            & 65536 \\
\hline
120             & 3,932,160        & 15,728,640       & 62,914,560       & 251,658,240 \\
\hline
480             & 15,728,640       & 62,914,560       & 251,658,240      & 1,006,632,960 \\
\hline
1920            & 62,914,560       & 251,658,240      & 1,006,632,960    & 4,026,531,840 \\
\hline
\end{tabular}
\caption{Total number of connections = 32x(Layers)x(Neurons) for different large sparse DNNs used in the Sparse DNN Challenge.}
\label{tab:RadiX-Net-DNNs}
\end{table}

\section{Input Data Set}

\begin{figure}[htb]
  	\centering
    	\includegraphics[width=\columnwidth]{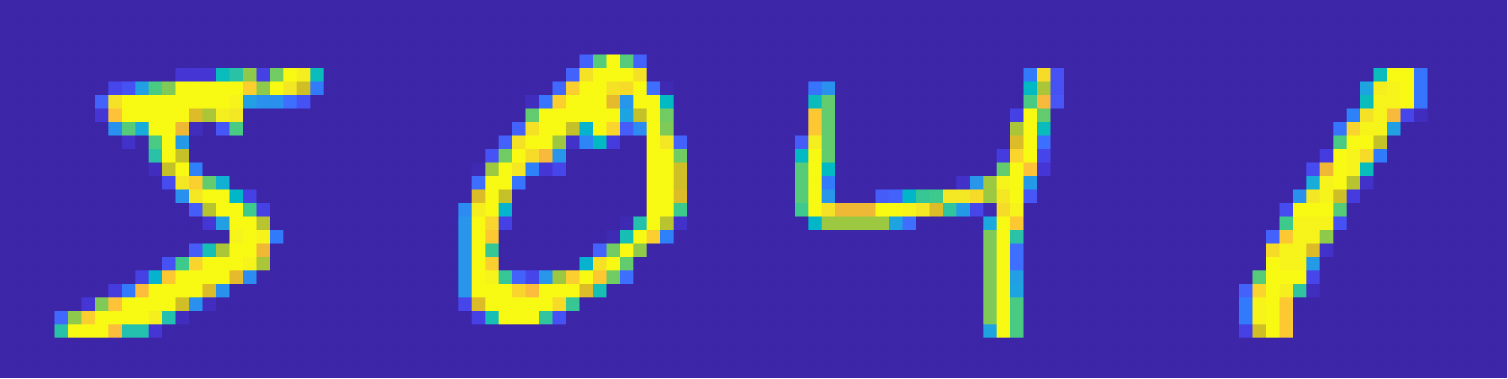}
	   	\includegraphics[width=\columnwidth]{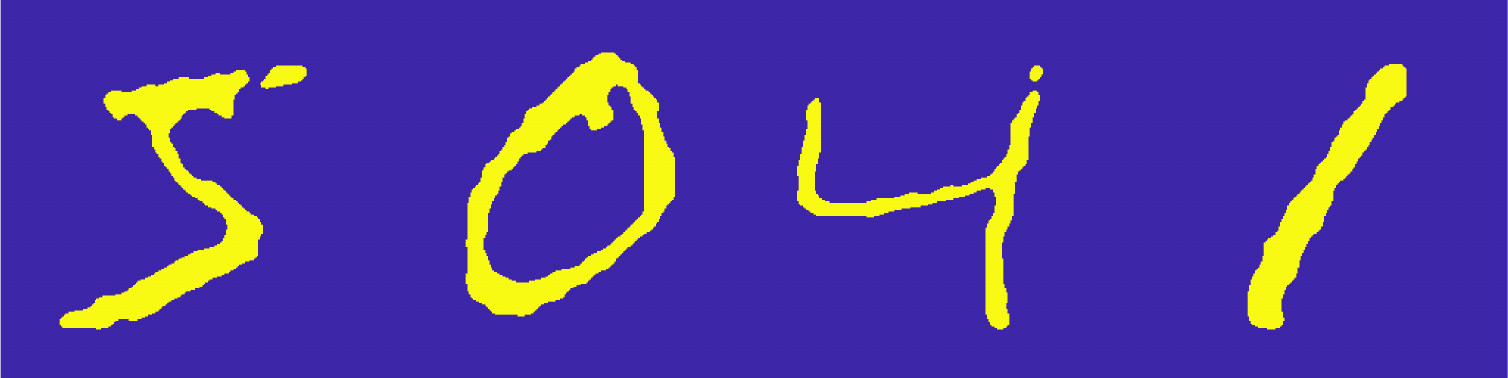}
	\caption{MNIST data set consists of 60,000 handwritten digits \cite{mnist}.  (top) Original 28{$\times$}28 pixel images of four MNIST images. (bottom) 256{$\times$}256 resampled thresholded versions of the same images.}
      	\label{fig:MNIST}
\end{figure}

Executing the Sparse DNN Challenge requires input data or feature vectors $\mathbf{Y}_0$.  MNIST (Modified National Institute of Standards and Technology) is a large database of handwritten digits that is widely used for training and testing DNN image processing systems \cite{mnist}.  MNIST consists of 60,000 28{$\times$}28 pixel images.   The Sparse DNN Challenge uses interpolated sparse versions of this entire corpus as input (Figure~\ref{fig:MNIST}).  Each 28{$\times$}28 pixel image is resized to 32{$\times$}32 (1024 neurons), 64{$\times$}64 (4096 neurons), 128{$\times$}128 (16384 neurons), and 256{$\times$}256 (65536 neurons).  The resized images are thresholded so that all values are either 0 or 1.  The images are flattened into a single row to form a feature vector.  The non-zero values are written as triples to a .tsv file where each row corresponds to a different image, each column is the non-zero pixel location and the value is 1.

\section{Sparse DNN Challenge}

The core of the Sparse DNN Challenge is timing DNN inference using the provided DNNs on the provided MNIST input data and verifying the output with the provided truth categories.  The complete process for performing the challenge consists of the following steps

\begin{itemize}
\item Download from GraphChallenge.org: DNN weight matrices $\mathbf{W}_\ell$, sparse MNIST input data $\mathbf{Y}_0$, and truth categories
\item Load a DNN and its corresponding input
\item Create and set the appropriate sized bias vectors $\mathbf{b}_\ell$ from the table
\item \underline{Timed}: Evaluate the DNN equation for all layers
$$
  {\bf Y}_{\ell + 1} = h({\bf Y}_\ell {\bf W}_\ell  + {\bf B}_\ell)
$$
\item \underline{Timed}: Identify the categories (rows) in final matrix with entries $> 0$
\item Compare computed categories with truth categories to check correctness
\item Compute rate for the DNN: (\# inputs) $\times$ (\# connections)  / time 
\item Report time and rate for each DNN measured
\end{itemize}

Submissions to the Sparse DNN Challenge are evaluated on the overall innovations highlighted by the implementation and two metrics: correctness and performance.  Correctness is evaluated by comparing the reported categories with the ground truth categories provided. The performance of the algorithm implementation is  reported in terms of the following metrics:
\begin{itemize}
\item Total number of non-zero connections in the given DNN: This measures the amount of data processed
\item Execution time: Total time required to perform DNN inference.
\item Rate: Measures the throughput of the implementation as the ratio of the number of inputs (e.g., number of MNIST images) times the number of connections in the DNN divided by the execution time.
\item Processor: Number and type of processors used in the computation.
\end{itemize}

\section{Community Submissions}

Graph Challenge has received a wide range of submissions across all its various challenges that have included hundreds of authors from over fifty organizations.  In 2019, twenty submissions across all the challenges were selected for publication \cite{Bisson-Nvidia-2019,Davis-TAMU-2019,Ellis-Sandia-2019,Wang-UCDavis-2019b,Wang-PingAn-2019,Mofrad-UPitt-2019,Pandey-Stevens-2019,Pearce-LLNL-2019,Acer-Sandia-2019,Yasar-GaTech-2019,Hoang-UTexas-2019,Wang-UCDavis-2019,Gui-HuazhongU-2019,Pearson-UIUC-2019,Blanco-CMU-2019,Ghosh-PNNL-2019,Liu-PNNL-2019,Almasri-UIUC-2019,Wanye-VaTech-2019,Huang-UIUC-2019}.  Six of the published submissions provided sparse DNN performance data for analysis \cite{Bisson-Nvidia-2019,Davis-TAMU-2019,Ellis-Sandia-2019,Wang-UCDavis-2019b,Wang-PingAn-2019,Mofrad-UPitt-2019}.

\begin{figure}[ht]
\centering
\includegraphics[width=2.4in]{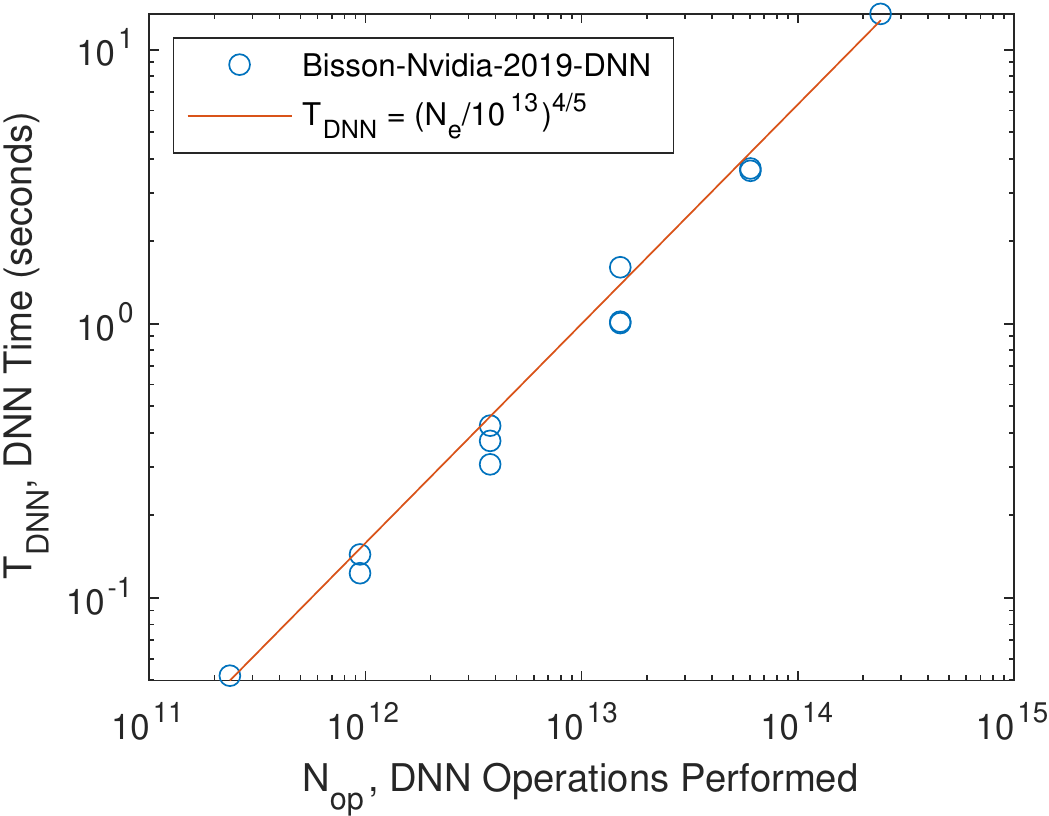}
\includegraphics[width=2.4in]{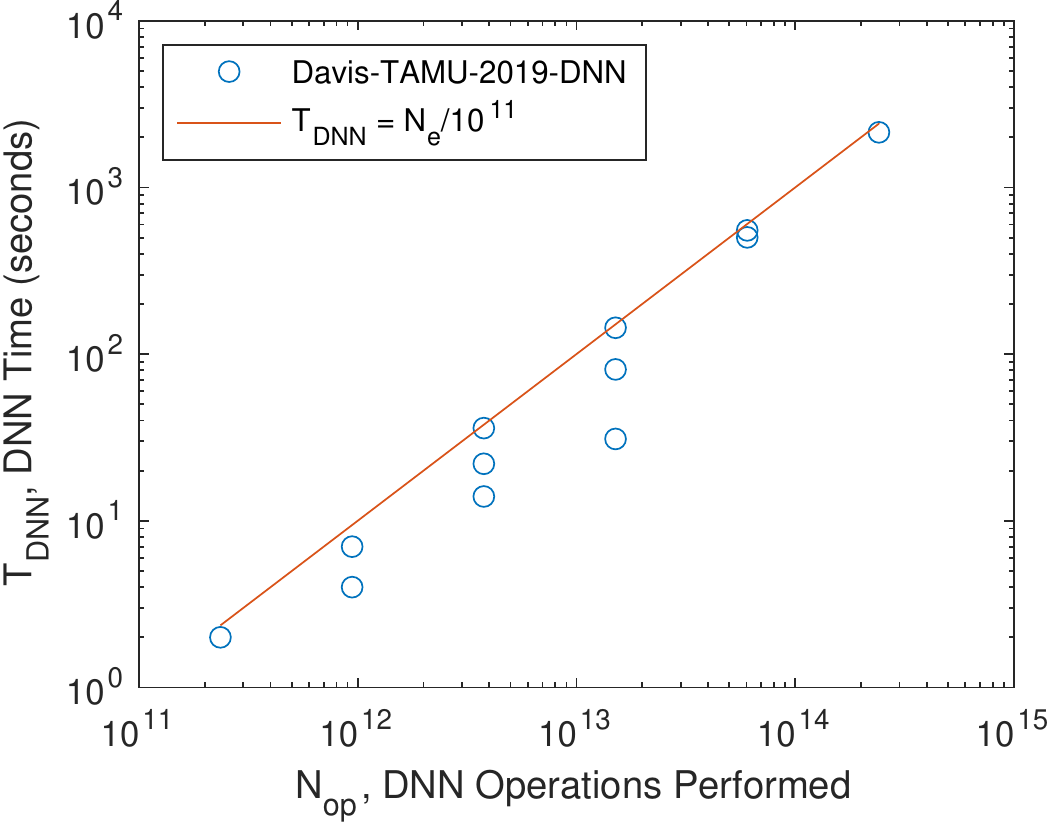}
\includegraphics[width=2.4in]{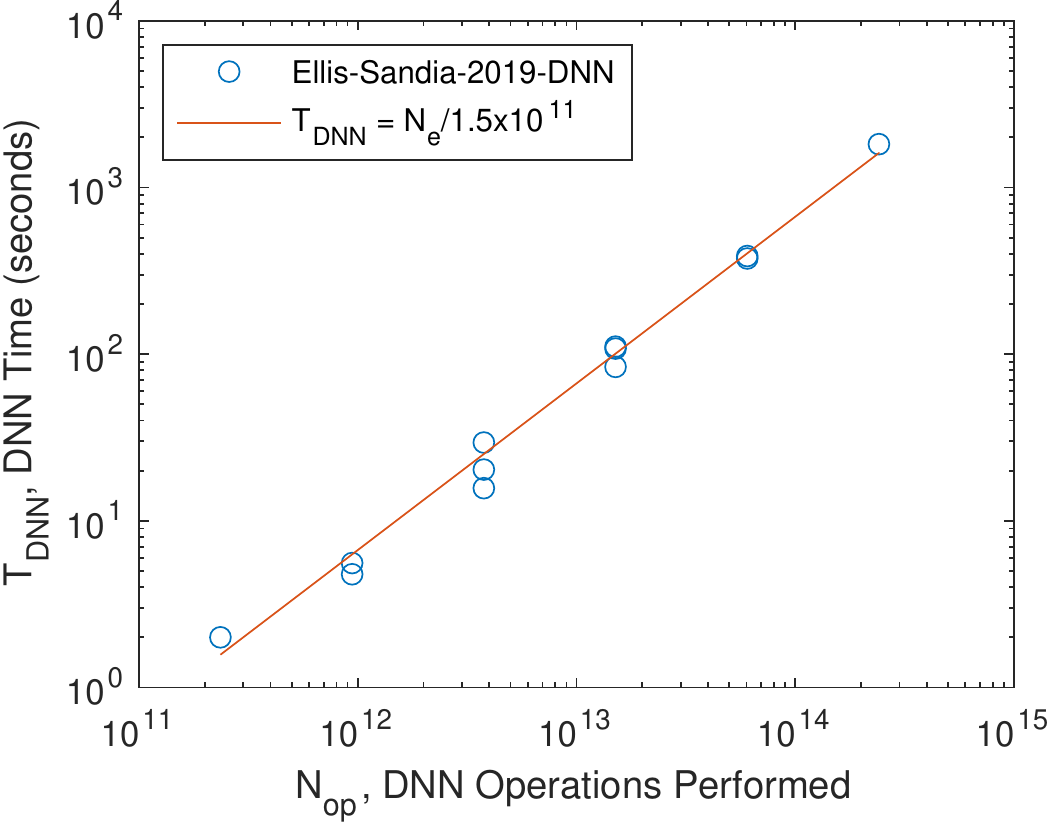}
\caption{2019 Champions and Innovation Award. Sparse DNN execution time vs number of operations and corresponding model fits for Bisson-Nvidia-2019 \cite{Bisson-Nvidia-2019}, Davis-TAMU-2019 \cite{Davis-TAMU-2019},  and Ellis-Sandia-2019 \cite{Ellis-Sandia-2019}.}
\label{fig:Champions}
\end{figure}

\begin{figure}[ht]
\centering
\includegraphics[width=2.4in]{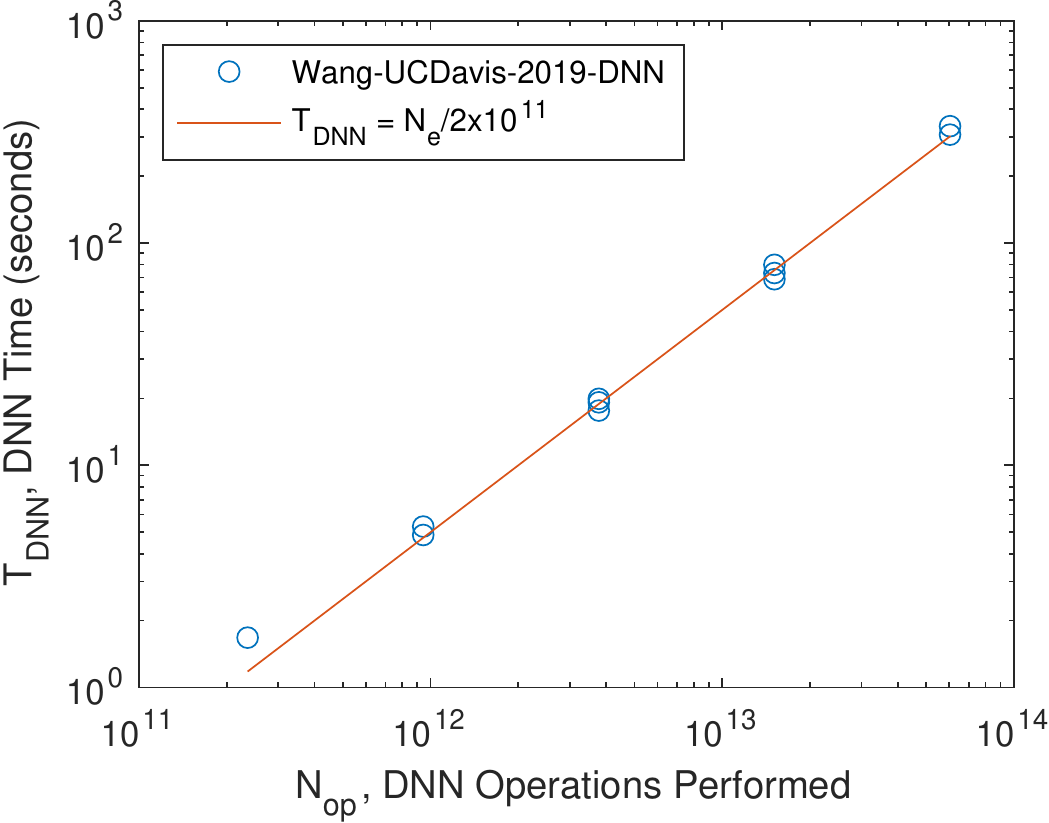}
\includegraphics[width=2.4in]{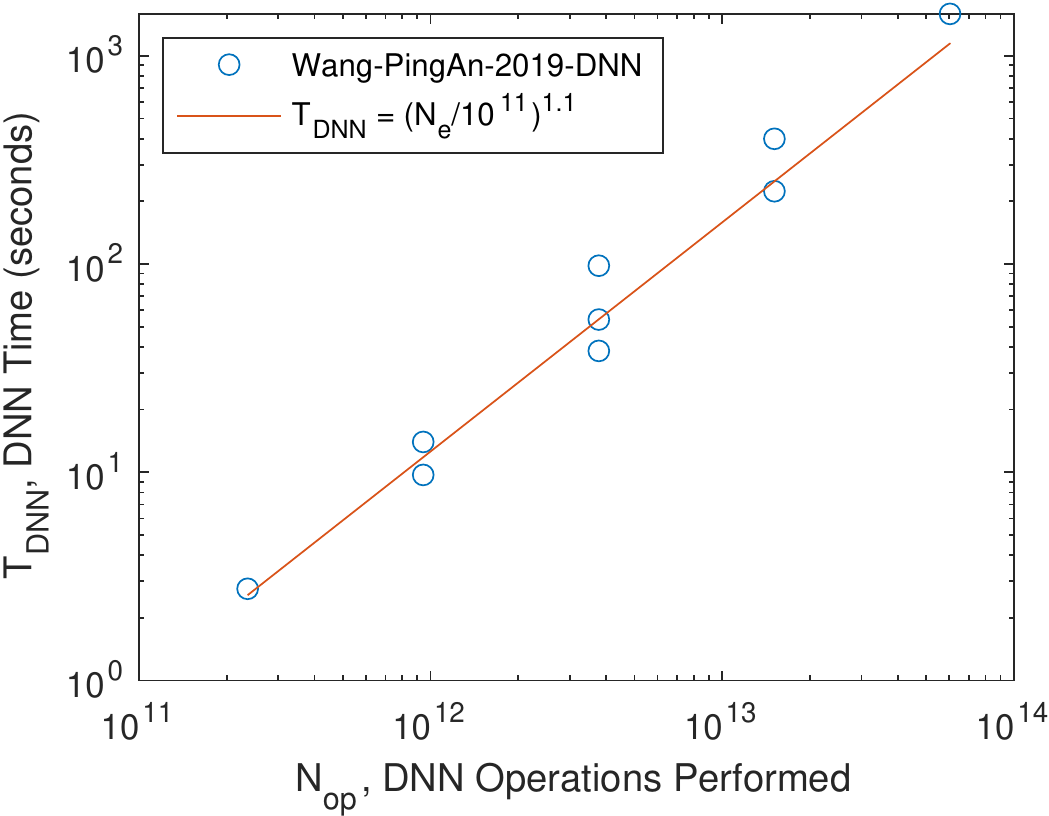}
\includegraphics[width=2.4in]{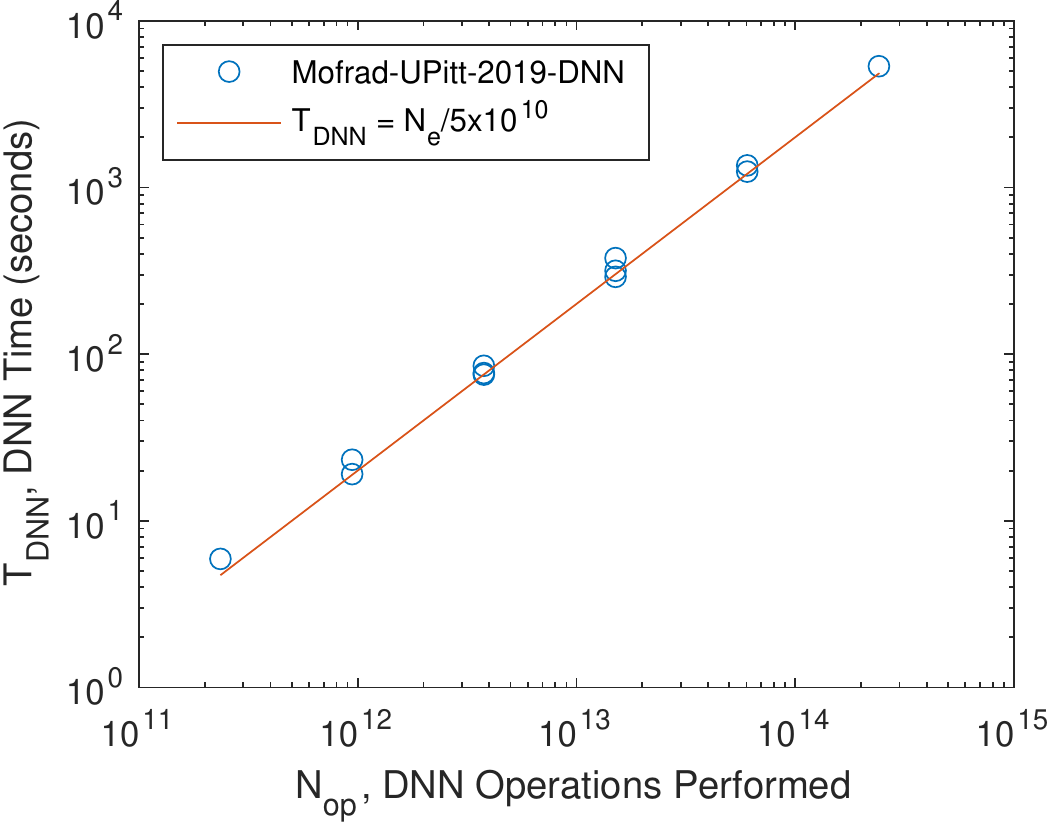}
\caption{Graph Challenge 2019 Student Innovation Award, Finalist, and Honorable Mention. Sparse DNN execution time vs number of operations and corresponding model fits for Wang-UCDavis-2019 \cite{Wang-UCDavis-2019b}, Wang-PingAn-2019 \cite{Wang-PingAn-2019}, and Mofrad-UPitt-2019 \cite{Mofrad-UPitt-2019}.}
\label{fig:Finalists}
\end{figure}

The submissions implemented the sparse DNN challenge in a comparable manner, resulting in over 60 distinct measurements of sparse DNN execution time, $T_{\rm DNN}$.  The number of connections, $N_{\rm c}$, in the graph describes the overall size of the graph.  The rate of operations processed in DNN inference is given by
$$
 {\rm Rate} = N_{\rm op}/T_{\rm DNN}
$$
where $N_{\rm op} = N_{\rm in} \times N_{\rm c} = 60,000 \times 32 \times L \times N$.  Analyzing and combining all the performance data from the submissions can be done by fitting a model to each submission and then comparing the models.  For each submission, $T_{\rm DNN}$ vs $N_{\rm op}$ is plotted on a log-log scale from which a  model can be fit to the data by estimating the parameters $N_1$ and $\beta$ in the formula
$$
   T_{\rm DNN} = (N_{\rm op}/N_1)^\beta
$$
where $N_1$ is the number operations that can be processed in 1 second. The sparse DNN execution time vs number of connections and corresponding model fits are shown in Figures~\ref{fig:Champions} and \ref{fig:Finalists}. The model fits illustrate the strong dependence of $T_{\rm DNN}$ on $N_{\rm op}$.   

\section{Performance Analysis}

\begin{figure}[ht]
\centering
\includegraphics[width=\columnwidth]{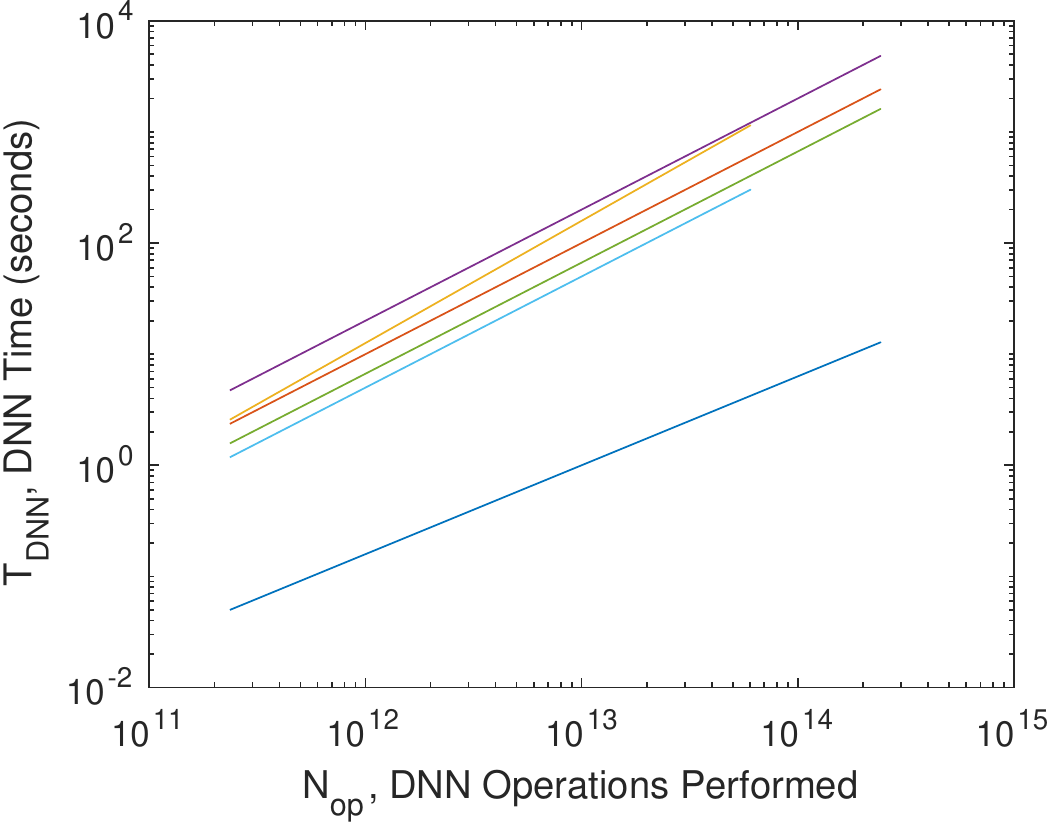}
\caption{Model fits of sparse DNN execution time vs number DNN operations for selected Graph Challenge 2019 sparse DNN  submissions.}
\label{fig:TimePerformance}
\end{figure}

\begin{figure}[ht]
\centering
\includegraphics[width=\columnwidth]{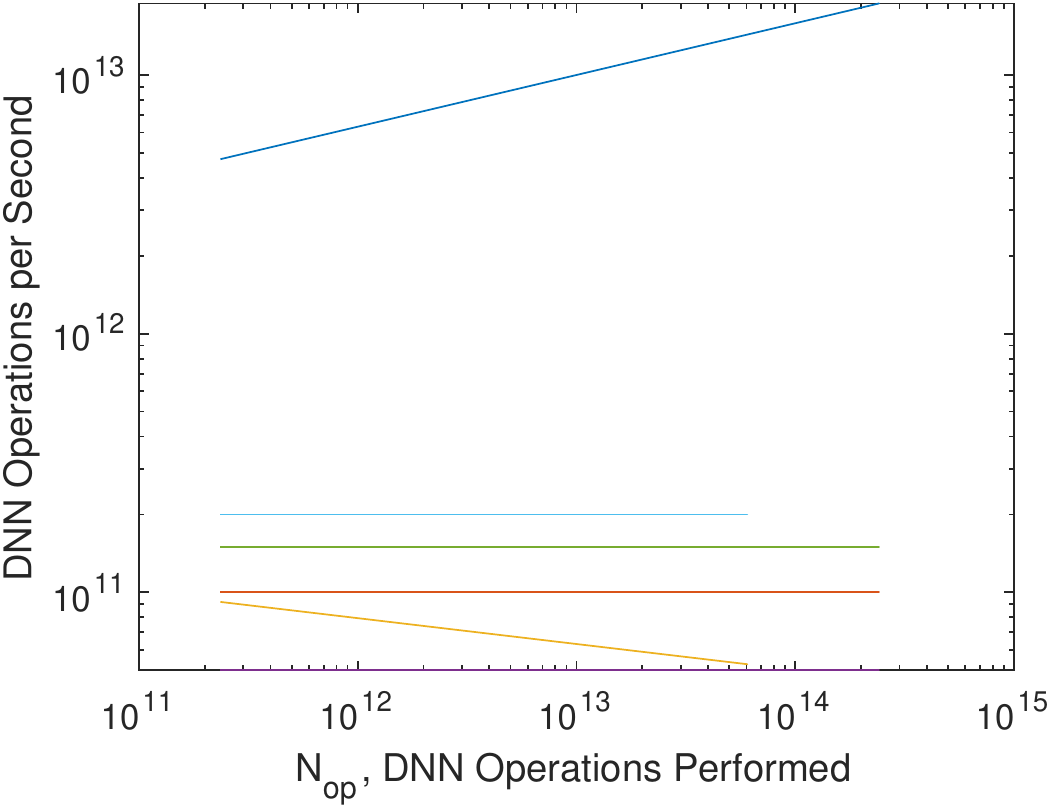}
\caption{Model fits of sparse DNN execution rate vs number DNN operations for selected Graph Challenge 2019 sparse DNN  submissions.}
\label{fig:RatePerformance}
\end{figure}

The normalized parameters $N_1$ and $\beta$, along with the largest values of $N_{\rm c}$, are shown in  Table~\ref{table:NormalizeModel2019} for each submission.  Submissions with larger $N_{\rm c}$, larger $N_1$, and smaller $\beta$ perform best.  The current state-of-the-art can be seen by plotting all the model fits $T_{\rm DNN}$ together (see Figures~\ref{fig:TimePerformance} and \ref{fig:RatePerformance}).   Combined, these suggest that typical performance model for 2019 is
$$
   T_{\rm DNN} \approx N_{\rm op}/10^{11}
$$
with the exception of \cite{Bisson-Nvidia-2019}, which produced the higher performance given by
$$
   T_{\rm DNN} \approx (N_{\rm op}/10^{13})^{4/5}
$$
Given that this is the first year of the sparse DNN challenge, it would be expected that subsequent submissions will aim to approach the higher performance demonstrated by \cite{Bisson-Nvidia-2019}.

\begin{table}
\caption{{\rm 2019 Sparse DNN time model fit coefficients for $T_{\rm DNN} = (N_e/N_1)^\beta$ for large values of  $N_e$.}}
\centering
\begin{tabular}{lllcc}
\hline
Ref & Submission & max $N_{\rm c}$ & $N_1$ & $\beta$ \\
\hline
\cite{Bisson-Nvidia-2019}      & Bisson-Nvidia-2019   & $4.0\times10^9$    & $1\times10^{13}$ & $4/5$ \\
\cite{Davis-TAMU-2019}         & Davis-TAMU-2019      & $4.0\times10^9$    & $1\times10^{11}$ & $1$ \\
\cite{Ellis-Sandia-2019}       & Ellis-Sandia-2019    & $4.0\times10^9$    & $1.5\times10^{11}$ & $1$ \\

\cite{Wang-UCDavis-2019b}      & Wang-UCDavis-2019    & $1.0\times10^9$    & $2\times10^{11}$ & $1$ \\
\cite{Wang-PingAn-2019}        & Wang-PingAn-2019     & $1.0\times10^9$    & $2\times10^{11}$ & $1.1$ \\
\cite{Mofrad-UPitt-2019}       & Mofrad-UPitt-2019    & $4.0\times10^9$    & $5\times10^{10}$ & $1$ \\
\hline
\end{tabular}
\label{table:NormalizeModel2019}
\end{table}

\section{Conclusion}

The MIT/IEEE/Amazon GraphChallenge.org encourages community approaches to developing new solutions for analyzing graphs and sparse data.  Sparse AI analytics presents unique scalability difficulties.  The machine learning, high performance computing, and visual analytics communities have wrestled with these difficulties for decades and developed methodologies for creating challenges to move these communities forward.  The  sparse Deep Neural Network challenge draws upon prior challenges from machine learning, high performance computing, and visual analytics to create a challenge that is reflective of emerging sparse AI systems.  The sparse DNN challenge is a based on a mathematically well-defined DNN inference  kernel and can be implemented in any programming environment. In 2019 several sparse DNN challenge submissions were received from a wide range of authors and organizations.   These submissions illustrate the state-of-the-art sparse DNN execution time, $T_{\rm DNN}$, is a strong function of the number of connections in the network, $N_{\rm c}$.


\section*{Acknowledgments}
%
%

The authors wish to acknowledge the following individuals for their contributions and support: Alan Edelman, Charles Leiserson, Steve Pritchard, Michael Wright, Bob Bond, Dave Martinez, Sterling Foster, Paul Burkhardt,  Victor Roytburd, Trung Tran, along with William Arcand, David Bestor, William Bergeron, Chansup Byun, Matthew Hubbell, Michael Houle, Anna Klein, Peter Michaleas, Lauren Milechin, Julie Mullen, Andrew Prout, Antonio Rosa, and Charles Yee.



\bibliographystyle{ieeetr}
\bibliography{aarabib}
%

\end{document}